# A Review of Affective Generation Models

Guangtao Nie, Yibing Zhan

**Abstract**—Affective computing is an emerging interdisciplinary field where computational systems are developed to analyze, recognize, and influence the affective states of a human. It can generally be divided into two subproblems: affective recognition and affective generation. Affective recognition has been extensively reviewed multiple times in the past decade. Affective generation, however, lacks a critical review. Therefore, we propose to provide a comprehensive review of affective generation models, as models are most commonly leveraged to affect others' emotional states. Affective computing has gained momentum in various fields and applications, thanks to the leap of machine learning, especially deep learning since 2015. With critical models introduced, this work is believed to benefit future research on affective generation. We concluded this work with a brief discussion on existing challenges.

—————————— ◆ ——————————

## 1 INTRODUCTION

Affective computing is an emerging interdisciplinary field where computational systems are developed to analyze, recognize, and influence affective states of a human. Affective computing (AC) involves multiple research disciplines, including psychology, neuroscience, cognition, computer science, and electrical engineering. AC can be generally divided into two subproblems: affective detection/recognition and affective generation [1], which are related to input signal processing and output expression generation of an intelligent agent, respectively.

With the popularization of mobile smartphone and wearable devices and the increase of data storage and processing capacity, AC in these years has gained momentum in various fields and applications, such as health, education, marketing, and advertising. Thanks to the recent leap of development in machine learning, especially the development of deep learning since 2015, publications on AC have soared. During the past period, affective detection/recognition have been reviewed several times. Nonetheless, there is a lack of literature review on affective generation. Consequently, we aim to deliver a comprehensive review of affective generation models since 2015.

Affective state is a notion that originated in psychology. It was first acknowledged in science in the 19th century [2], [3], [4]. At first, emotion was considered irrelevant with rigorous scientific research and even as an obstacle for rational decision making and reasoning. Damasio refuted the above bias towards emotion by acknowledging its positive role in rational thought: the combined functionality of of emotion and body contributed to rational human thinking [5]. In this work, affective state and emotional state are used interchangeably as in most literature [1], [6], [7].

AC, in contrast, has not been proposed and drew public attention until 1990s, with remarkable works from Picard and Scherer [1], [8]. It was first described as computing that related to, arose from, or influenced emotions [1]. Several works have tried to categorize the research topics under AC, but there has not been any consensus yet [9]: Tao and Tan divided AC into three general topics: affective understanding, affective generation, and application [10]. Carberry and de Rosis divided AC into four areas, namely, the analysis and characterization of affective states, automatic recognition of affective state, adaption of response to user's affective state, and expression and exihibition of affective state [11]. In this work, we adapted the category proposed by Tao and Tan, with affective understanding renamed as affective detection/recognition. We suspect the capability of current technology to really understand emotions as a human does, and we do not include applications in our work, since applications are not the key of this work.

### 1.1 Related works

There have been several works of literature review/overview/summary on AC, but few explicitly concerned the affective generation:

Back in 2005, Tao and Tan reviewed AC development, with focuses on emotional speech processing, facial expression analysis and synthesis, body gesture and movement analysis, multi-channel information processing, and affect understanding and cognition [10]. However, they did not provide a comprehensive review on affective text/speech and movement synthesis.

Calvo and Mello [6] delivered a survey on affect detection on the first issue of IEEE Transactions on AC back in 2010. Emotion theories, affect detection methods, and input signals ranging from facial expressions, voice, body language and posture, physiology, brain imaging, text, and their multi-channel fusion were reviewed and discussed. The survey [6] was a landmark in AC, even though it did not cover affective generation.

Published in 2017, Poria et al. [12] reviewed affect recognition from the perspective of multimodal information fusion. They reviewed multimodal affect analysis and recognition framework based on the input of audio, visual, and text information. Public datasets and off-the-shelf APIs re-

————————————————

- *G.N. is with JD.com, China. E-mail: nieguangtao1@jd.com.*
- *Y.Z. is with with JD Explore Academy, China. Email: zhanyibing@jd.com.*

lated to affect recognition were also summarized. Nonetheless, they only focused on affect recognition, and they did not include physiology information as input.

The latest related work came from Arya et al. [13]. They surveyed three critical questions related to AC: the contribution from interrelated domains to AC, prominent applications of AC, and research challenges and the pertinent issues. Their survey was conducted through the perspective of mutil-disciplinary contribution such as physiology, psychology, computer science, sociology, mathematics, linguistics, and their fusion. However, they did not mention state-of-the-art technologies applied, neither did they review affective generation techniques.

Besides the preceding general-purpose review of AC, back in 2013 [14] and 2019 [15], expressive body movement generation for robots were surveyed. Other works reviewed: 1) AC in specific applications such as education [16], [17], medicine [18], [19], workplace [20], human machine interaction [21], and video gaming [22]; 2) AC on certain media including mobile devices [23], [24], [25] and multimedia [26]; 3) AC with certain signal input or influence factors, like EEG [27], [28], [29], color and temperature [30], [31], [21], gender and age [32]; 4) and others [33], [34].

In this work, we aimed to deliver a general-purpose literature review on affective generation models. Specifically, modalities of affective generation of text, audio, facial expression, and body movement are covered.

### 1.2 Criteria for inclusion in the literature review

Extensive literature reviews on affective generation ever since 2015 are covered in this work, as this is the period when deep learning [35] booms and benefits AC significantly. Moreover, only models related to affective generation are covered, topics such as hardwares, sensors, ethics, human computer interaction, and human robot interaction will not be discussed.

### 1.3 Structure

The remainder of this paper is organized as follows: the background of emotion theories and models are introduced in Section 2. Affective generation models in different areas are summarized in Section 3. Discussions on challenges and future works are delivered in Section 4. Finally, this work is concluded in Section 5.

## 2 EMOTION

### 2.1 Emotion theories

First of all, there is no consensus on the definition of emotion up till now [36], [37]. The history of emotion theories is long and diverged [23]. Here we introduce three types of emotion theories, namely, psychology, cognition, and neurobiology.

Several theories were developed for the psychology of emotion, namely, introspective, behavioral, and psychoanalytic [4]. The introspective theory focuses on the feeling states and relating visceral physiology to feelings [4], [38], [39]. The behavioral theory can date back to Darwin, who claimed that all men, including some animals, express emotions through similar behaviors [2]. The psychoanalytic theory was developed by Freud and Breuer, who believed that emotion could only be inferred through human verbal reports, expressive behavior, or dreams [40].

There is also cognition of emotion [41]. In a cognitive appraisal theory, it is hypothesized that emotion coordinates quasi-autonomous processes in the nervous system. It helps to maintain and accomplish transitions between plans and in systems with multiple goals [42].

Besides psychology and cognition, neurobiological theories were also proposed. In facial-feedback theory, it is believed that skeletal muscle feedback from facial expression plays a causal role in regulating emotional experience and behavior [43]. Damasio hypothesized that brain responses constitute the (body expression of) emotion and emotion feeling can be regarded as a consequence of neurobiological expression [44]. Izard, nonetheless, proposed that emotion is a phase, rather than a consequence of neurobiological activity or body expression of emotion [45], [46].

For more detailed discussions of emotion theories, readers can refer to [1], [45], [47], [48], [49], [50], as this is not the main focus of this work.

### 2.2 Emotion models

Numerous models have been proposed to model the relationships between different emotions. These models can be approximately classified into two categories: dimensional and discrete emotion approaches [51].

Dimensional models are proposed based on the hypothesis that the core effect of emotion is inherently continuous. A dimensional model provides important information such as valence and arousal of the stimuli [45], [52]. In dimensional models, emotions are projected onto several dimensions [48], [53], including valence, arousal, and dominance. Valence is usually used interchangeably with pleasure, with negative as unpleasant and positive as pleasant. Arousal is sometimes in lieu of excitement or activation, which represents how strong emotion is. Dominance stands for to what extent a participant is in control of the emotional state.

Some dimensional models, such as PAD and SAM, involve three dimensions. Pleasure-arousal-dominance (PAD) model is a three-dimensional emotion model [54]. The three axes formed a triangular coordinate system, where an emotional state can be projected. The Self-Assessment Manikin (SAM) model was proposed for self-evaluation on valence, arousal, and dominance [55]. There are five manikins in each of three rows, representing valence from positive to negative, arousal from high to low, and dominance from minimum control to maximum control, respectively. In the first row, a manikin with a sad face represents negative valence, while one with a smiling face represents positive valence. In the second row, a manikin with large excitement represents high arousal, and one with small excitement stands for low arousal. In the third row, small manikin indicates minimum control, whereas large manikin denotes maximum control.

The majority of dimensional models typically focuses on valence and arousal:

FEELTRACE [56] is a dimensional model where the most intense emotions define a circle. The horizontal axis ranges from very negative to very positive while the vertical axis ranges from very passive to very positive. It is designed for observers to describe perceived emotional state by moving a pointer to the appropriate point in the model. Circumplex model is a two-dimensional plane where valence and arousal/activation are two dimensions [57]. Valence is the horizontal axis, which ranges from unpleasant (negative) to pleasant (positive), while arousal/activation is the vertical axis ranging from deactivation (negative) to activation (positive). The two axes cross at the neural state. The vector model can be considered as a rotation of Circumplex model, 90 degrees counterclockwise [58]. The positive and negative affect schedule (PANAS) model provides 10-item scales for both positive affect and negative affect [59]. It can be used with both long-term and short-term labelling of emotions. The Activation-Deactivation Adjective Check List (AD ACL) is a two-dimensional self-rating test constructed and extensively validated for rapid assessments of momentary activation or arousal states [60]. Energetic arousal and tense arousal are the two core dimensions.

The aforementioned dimensional models emphasize the conscious experience or phenomenology of affect. In contrast, there are also hypotheses of the existence of a discrete emotion or pattern of interaction emotion in the conscious brain [61], [62], [63]. Discrete emotion models provide several basic discrete emotion, on which complex emotions are believed to form [51]. Differed by the number of basic emotions, there have been numerous types of discrete emotion models [64]:

Paul Ekman proposed a basic emotion model, of which the six emotions were believed to be culturally universal [65], [66]: anger, disgust, fear, happiness, sadness, and surprise [67]. Plutchik proposed eight primary emotions: fear, anger, joy, sadness, acceptance, disgust, expectancy, and surprise. Arranged in a circle in the form of "emotion solid", primary emotions can form more complicated emotions in the third dimension based on intensity. Nonetheless, the plausibility of the "emotion solid" assumption was questionable. In contrast, Oatley and Johnson-Laird made no assumption and treated five basic emotions (happiness, anger, anxiety, sadness, and disgust) as separate categories. Complex emotions can also be constructed based on these five emotions. For example, "Differential Emotions Theory" was proposed by Izard, which consisted of 10 basic emotions (interest, enjoyment, surprise, sadness, anger, disgust, contempt, fear, shame, and shyness) [68]. In this model, each state had respective substate, unique organization, motivational features, facial expression, and evolutionary basis [51]. Profile of Mood States (POMS) is another discrete emotion model that consists of 65 items for evaluating six basic moods: anger, confusion, depression, fatigue, tension, and vigor [69].

The discrete emotion models have drawn criticism for their varying number of basic emotions in different models [64]. Some of the basic emotions adopted in the aforementioned models, such as interest, surprise, acceptance, and expectancy, were even being questioned for their qualification as emotion [51].

Besides dimensional and discrete emotion approaches, hybrid models are developed based on the combination of dimensional and discrete emotion approaches: Smith and Ellsworth proposed eight cognitive appraisal dimensions to differentiate emotional experience [70]. In their study, subjects were asked to recall past experience with each of 15 emotions and rate them along 6 dimensions, pleasantness, anticipated effort, certainty, attentional activity, self-other responsibility/control, and situational control. In addition, Diener et al. proposed a hybrid model with structural equation modelling: six basic emotions (love, joy, anger, fear, shame, sadness) were combined with negative-positive valence dimension [51], [71].

## 3 AFFECTIVE GENERATION MODELS

Affective generation can be described as generating messages to influence the emotional states of users. In Picard's description, affective generation application is a computational system that expresses what a human would perceive as an emotion [6].

In affective detection/recognition for an intelligent agent, the inputs include textual, audio, visual, bodily, facial, and physiological modalities. In correspondence, the outputs consist of all modalities except physiology. In this section, affective generation is reviewed in five categories: affective text generation, affective audio generation, affective facial expression generation, and affective movement generation.

Generation techniques are applied either for virtual agents, or embodied agents such as robots and embodied conversational agents.

### 3.1 Affective text generation

We combined affective text generation and affective dialogue generation, as dialogue generation can be viewed as back and forth text generation, with text as input [72]. There are two types of text generation algorithms, namely, text-to-text generation and data-to-text generation [72].

It is widely acknowledged that the following six sub-problems constitute natural language generation: *content determination, text structuring, sentence aggregation, lexicalization, referring expression generation, and linguistic realization* [72], [73]. The effort of generating affective text has been predominantly put on *lexicalization* [74], [75], [76], where models were proposed to add affect intensity onto generated text/dialogue without losing grammatical correctness.

Generating emotional language is a key step towards building empathetic natural language processing agents [77]. Having the capability to generate and express affect during human-machine interaction is one of the major milestones in machine learning development [78]. Studies have shown that addressing affect in dialogue systems can enhance user satisfaction [79], leading to positive perception [80] and fewer breakdowns [81] in dialogues.

Agents endowed with affective text-based dialogue generation system are generating interest both in academy [82] and the industry. Similar to the advancement of machine learning, affective text generation has also gone through the evolution from hand engineered features [83], [84] to automated feature extraction [74], [76], [85], [86], [87], [88], [89].

Most works on affective dialogue generation are based on enconder-decoder architectures, such as Sequence to Sequence (Seq2Seq) [90] text generation model. Encoder-decoder architectures are frequently used for machine translation and free form question answering.

Song et al. proposed an emotional dialogue system (EmoDS) that can generate meaningful responses with a coherent structure for a post, and meanwhile express the desired emotion explicitly or implicitly within a unified framework [91]. The EmoDS consisted of encoder-decoder architecture with lexicon-based attention mechanism. An emotion classifier provided global guidance on the emotional response generation by increasing the intensity of emotional expression. Zhong et al. proposed an end-to-end affect-rich open-domain neural conversational model, Affect-rich Seq2Seq (AR-S2S), which produced affective responses appropriate in syntax and semantics [78]. The proposed AR-S2S extended Seq2Seq model with VAD (valence, arousal, dominance) embedding and affective attention. Emotional Chatting Machine (ECM) [86] was an affective conversational model based on Seq2Seq model. It modeled high-level abstraction of emotion expressions by embedding emotion categories. The change of implicit emotion state was captured by an internal memory while external memory helped generate explicit emotional words. In Mojitalk [77], authors applied several state-of-the-art neural models, such as Seq2Seq, Conditional Variational Autoencoder (CVAE), and Reinforced CVAE, to learn a generation system that was capable of responding with an arbitrary emotion. They also trained a large-scale emoji classifier and validated its accuracy on the generated responses.

In models including ECM, users need to input the desired emotion [77], [86], [91], [92], [93], [94], [95], whereas for the model proposed by Asghar [74], they modeled emotion with affective word embedding as input. Asghar et al. [74] proposed an affective text generation model based on Seq2Seq. They used cognitive embedding of words by affective dictionary. Moreover, affective objective was added onto the cross-entropy loss function, and affectively diverse beam search algorithm was adopted to inject affective diversity.

Besides the aforementioned works, there are other works investigating special topics or conditional problems: Valitutti and Veale investigated the way of inducing ironic effects in automated tweets [96]. They studied two types of irony: The first type is adjectives with opposite polarity and contrastive comparison, used to produce ironic incongruity. The second type is irony markers, including scare quotes and hashtag #*irony*. Valence, irony, surprise, humor, and retweetability of the automated tweets were investigated in their experiments. Ghosh et al. proposed Affect-Long Short Term Memory (Affect-LM) to generate affective conversational text based on context words, affective category, and affective strength [85]. They explicitly used two parameters e and β, both manually adjustable and automatically inferrable, to represent the affective category and strength, respectively.

Affective dialogue generation can be divided into two subproblems based on the application scenarios: one is virtual agent, such as Siri, which delivers information through voice; the other is embodied agent, such as embodied conversational agent, which conveys both verbal and non-verbal information [97].

Affective image captioning can be viewed as a problem under the category of data-to-text generation, with image as input: Yang et al. [98] proposed a method to generate captions with distinctiveness and attractiveness, which consisted of a content module and a linguistic module. The content module mapped the image into deep representations, where emotion was classified. The linguistic model took these deep representations as input, then embedded the training results into latent space, where frequency vector was clustered as output. The outputs from both content and linguistic module were integrated before inputting into a LSTM module for affective caption generation. Affective Guiding and Selective Attention Mechanism (AG-SAM) [99] was also proposed for affective image captioning. Deep representations were first extracted from input images by CNN, followed by paralleled attention gate and affective vector. The attention gate decided the amount of visual information fed into a LSTM module and the affective vector was trained based on deep representations to reflect emotional response, which was then input into the LSTM module. Finally, the LSTM module worked to generate affective captions.

### 3.2 Affective audio generation

Affective text/dialogue generation should be combined with affective pronunciations, as the utterance of the synthesized speech is another important issue that can influence the affective performance of an agent [100].

Both affective speech synthesis and affective music synthesis are covered in this subsection.

#### 3.2.1 Affective speech synthesis

There are two types of affective speech synthesis models: one is affective Text-to-Speech, where affective audio spectrogram is synthesized directly from text; the other is adding emotional component onto neutral speech. All related works considered improving the naturalness of synthesized affective speech, which was mostly validated through subjective ratings. Different objective measurements were also included in some works, depending on the concrete problem background.

It has been more and more prevalent for Text-to-Speech models synthesizing natural and affect rich speech, instead of machine-like sound. Typically, the affective Text-to-Speech synthesis can be divided into two subproblems, namely, detection of affect in text [101], and generation of speech in line with the affective message [100].

There are several works on affective speech synthesis based on the modification of Tacotron [102], an end-to-end speech synthesizer based on the Seq2Seq model with attention paradigm, which took texts as input and output audio spectrogram. The generated speech can then be fed into Griffin-Lim reconstruction algorithm [103] to synthesize speech clips: Lee et al. introduced context vector and residual connection at RNN to address the problem of exposure bias and irregularity of attention alignment [104]. Li et al. [105] proposed a controllable emotion speech synthesis approach based on emotion embedding space learned from references. In order to deliver the emotion more accurately and expressively with strength control, they modify the Prosody-Tacotron structure with two emotion classifiers to enhance the discriminative emotion ability of the emotion embedding and the predicted mel-spectrum.

Other works on Text-to-Speech demonstrated the importance of features and the superiority of models: Tahon et al. [100] proposed double adaption to improve the quality and expressivity of the synthesized sentences. Double adaption refers to adapting voice and emotional pronunciation. Their work exhibited the importance of prosodic features. An et al. [102] investigated the performance of two lstm-rnn based models, one as emotion dependent model, the other as an unified approach with emotion as code. The two models both outperform traditional HMM based models [103] in terms of synthesized utterance naturalness.

Choi et al. [104] considered the scenario of multi-speaker emotional speech synthesis, they used emotion code and mel-frequency spectrogram as emotion identity. Trainable speaker representation and speaker code were used for speaker variation. Objective measures such as mel cepstral distortion, band aperiodicity distortion, fundamental frequency distortion in the root mean squared error, and voice/unvoiced error rate, along with subjective measures on naturalness, speaker similarity, and emotion similarity, were provided as validation for their proposed method's superiority.

Huang et al. [105] presented a work of emotional 3D talking head synthesis. A temporal restricted Boltzmann machine was adopted for emotion transfer between neutral faces and emotional ones while an LSTM-RNN was used for emotional voice conversion from neutral one. Four discrete emotions were realized, and human raters were invited to classify the synthesized results.

In [106], [107], it was revealed that duration and timing were two critical issues in the manual synchronization of gesture with speech for conversational agents.

Adding emotional components onto neutral speech has been repeatedly investigated using statistical methods, which are called statistical parametric speech synthesis (SPSS): In [108], emotion additive model (EAM) was introduced to represent the difference between emotional and neutral voices. EAMs can be linearly combined to create certain emotional speech. Eigenvoice was also constructed for better synthesis in target emotion speech. In [109], Xue et al. studied the model adaption models in this problem: retraining model with emotion-specific data, augmenting model input using emotion specific codes, and using emotion depedent output layers with shared hidden layers. An et al. [110] proposed to combine a LSTM with a mixture density network for multimodal regression and variance prediction. KL-divergence was also introduced to maximize the distance between the distributions of emotional speech and neutural speech. Inoue et al. [111] also investigated this problem with several models: the parallel model, the serial model, and the auxiliary input model.

### 3.2.2 Affective music synthesis

It is suggested that music generation approaches can be divided into two categories: transformational and generative algorithms [115]. Affective music synthesis models all fall within the latter category, which needs to create the musical structures, as opposed to acting on prepared structure [116]. Some models take visual information as input, some composite music for computer games related to dynamic environments, and others take advantage of human physiology or communication.

Sergio and Lee [112], [113] designed RNNs and Neuro-Fuzzy Networks to generate music that aimed to convey similar emotions as the input image, by trained with the image-music pairs extracted from video excerpts. Viewers were invited to rate the input images as either positive or negative emotions. The Neuro-Fuzzy Network was adopted to classify the binary emotion of input images, with viewers' ratings as labels in the training stage. Two RNNs were designed for music generation, following the prediction of Neuro-Fuzzy Network, one for positive emotion and the other for negative emotion. They later developed Scene2Wav [114], which was dedicated for the same problem, with improved neural network architecture and synthesis results.

Scirea et al. [115] proposed MetaCompose for affective music composition in dynamic environments, such as computer games. MetaCompose consisted of a graph traversal-based chord sequence generator, a search-based melody generator, a pattern-based accompaniment generator. In particular, they investigated the problem of diverse solutions with multi-objective optimization. Viewers were recruited to rate the synthesized music and its expressed valence.

Other affective music composition works include autonomous agent-assisted affective music generation [116], where users were supposed to communicate with an autonomous agent about their emotion before it composed music with the desired emotion preference. The agent was programmed with the basic compositional rule of tonal music. Affective music generation based on brain-computer interfaces, such as electroencephalogram (EEG), was also investigated, where emotion features were extracted from EEG signal, which then directed the composition of emotional music [117]. The problem of long-term coherence when generating affective music was investigated by MorpheuS [118].

### 3.3 Affective facial expression generation

Based on the characteristics of input/output images, facial expression generation can be categorized into virtual avatar animation and photorealistic face synthesis [119]. Depending on whether the input and the output shared the same identity or not, facial expression generation could be categorized into facial expression editing (FEE) and facial expression transfer (FET). In FEE, the facial expression is changed on a given portrait, whereas in FET, the model works to transfer facial expression between different iden-tities. Based on the adopted models, facial image expression syntheses can be categorized into traditional graphic-based models and emerging generative, data-driven approaches. All methods either rely on the extraction of features from geometric [120], appearance [121], RGBD space [122], [123], or crossfading, wrapping/morphing existing faces [119].

The following work is an example of FEE adopting a graphic-based model: Kollias et al. [124] proposed a model that took as input an arbitrary face with neutral expression and synthesized a new expression on it with either a target emotion category or emotion valence and arousal. The model fit a 3D morphable model on an input image, then deformed the reconstructed face and added the input affect, finally blended the new face with the given affect into the original image.

By contrast, more and more works adopted data-driven approaches, mostly resorting to deep neural networks, such as generative adversarial networks (GANs) [125], conditional GANs (cGANs) [126], and conditional variational autoencoders (cVAEs) [127], which required large scale training data instead of paired samples, as opposed to some of the graphic methods, to properly disambiguate identity information [124].

Zhou proposed a conditional difference adversarial autoencoder (CDAAE) for photorealistic FET [119]. The CDAAE learned to generate an unseen person's facial expression with a discrete target emotion or facial action unit. The CDAAE worked partially due to the newly added feed-forward path connecting low-level features at the encoder with the corresponding level at the decoder. In [128], ExprGen was presented to take as input images of human faces and generate the character rig parameters accordingly for virtual avatar animation. ExprGen was a multi-stage deep learning system using the latent variables of human and character recognition convolutional nets to control 3D animated character rig. The multi-stage deep learning system consisted of 3D-CNN and Character Multi-Layer Perceptron. The multi-stage deep learning system was trained using five publicly available labeled facial expression datasets.

Instead of using discrete emotion labels, continuous labels/dimensions were also adopted in some works: Ding et al. proposed Expression GAN (ExprGAN) [129] for photorealistic FEE with controllable expression intensity, which enabled the expression intensity to be continuously adjusted from low to high. Tang et al. [130] tried fine-grained expression manipulation with expression-guided GAN (EGGAN), which could synthesize photorealistic images with continuous intermediate expressions based on continuous emotion labels and structured latent codes. Pham et al. [131] presented generative adversarial talking head (GATH), a deep neural network that enabled facial expression synthesis of a given portrait with continuous action unit coefficients. Their model directly manipulated image pixels to generate various expressions on an unseen face, while maintaining features such as facial geometry, skin color, hair style, and surrounding background. In [132], GANimation was introduced as a novel GAN that could control the magnitude of activation of each action unit and several combinations of them.

There are also works trying to combine the strengths of graphic methods and data-driven approaches:

A Geometry-Guided Generative Adversarial Network (G2-GAN) was designed for continuously adjustable and identity preservable facial expression synthesis (FEE) [133], where facial geometry was adopted to guide the generation of facial texture synthesis with a certain expression. In this model, paired submodels were jointly trained for expression removal and expression synthesis, which formed a cycle between neutral expression and any other expression. Similarly, a Geometry-Contrastive Generative Adversarial Network (GC-GAN) [134] was proposed for transferring continuous emotions across different subjects. The embedded geometry is injected into the latent space of GC-GAN as continuous conditions to guide the generation of facial expressions effectively. Yeh et al. [135] presented an automatic approach of FEE, such as from smiling to neutral. Their approach combined flow-based face manipulation with the generative capability of Variational Autoencoder (VAE), which learned to encode the flows among different expressions in a latent space. Their results demonstrated higher perception quality than previous methods adopting VAEs.

More than developing affective facial expression generation models on images, Thies et al. [136] further tried face reenactment, which did facial expression synthesis in videos: Their model animated the facial expressions of a target video by a source actor and re-render the manipulated output video in a photorealistic fashion. Technically, the model first dealt with facial identity recovery by non-rigid model-based bundling. The model then tracked facial expressions of both target and source videos by a dense photometric consistency measure. The video reenactment was finally achieved by deformation transfer between source and target [136]. They later added other functionalities, extending the modelling of facial expressions onto head, eye, and kinematic torso [137]. ReenactGAN [138] was engineered to transfer facial movements and expressions from an arbitrary person's monocular video input to a target person's video. It mapped the source face to a boundary latent space, adapted source face's boundary to the target's boundary, and generated the reenacted target face. Ma and Deng [139] proposed a real-time end-to-end facial reenactment system, without the need for any driving source. It could generate desired photorealistic facial expressions on top of input RGB video, with an unpaired learning framework developed to learn the mapping between any two facial expressions in the facial blendshape space. Otberdout et al. [140] proposed a facial reenactment model, exploiting the face geometry by modelling the facial landmarks motion on a hypersphere. A GAN was adopted to generate facial landmark motion in the hypersphere to synthesize various facial expressions.

Beyond facial expression synthesis on images and video reenactment, there were also works on affective image-to-video translation, a FEE problem: Conditional MultiMode Network (CMM-Net) [141] was devised for generating various facial expression videos with distinctive characteristics, given a neutral face image and a discrete emotion label.

The input face image and emotion label were used to generate landmark sequences, which were used to guide the translation from the neutral image into facial expression video. Potamias et al. [142] advanced the image-to-video translation problem from 2D to 3D: they took as input 3D meshes instead of 2D images and, therefore, their generative model output 4D facial expression video. Their experimental results demonstrated the preservation of identity and high-quality expression synthesis. Similarly, Motion3DGAN [143] exploited a set of 3D landmarks to generate expressive 4D faces by modeling the temporal dynamics of facial expressions using a manifold-valued GAN, with a neutral face as input.

Other works include FET-FEE fusion and talking-head video generation with affect rich facial expression:

In 2019, Ali and Hughes [144] introduced Transfer-Editing and Recognition Generative Adversarial Network (TER-GAN), a model to realize three functions: facial expression transfer, editing, and recognition. When doing FET in TER-GAN, two encoders were adopted to encode identity and expression information from the source and target input image, respectively. When doing FEE in TER-GAN, two portraits with the same identity but different facial expressions were input to the two encoders. The generated images, either in FEE or FET, were more blurry, compared with ones generated by ExprGAN. Zeng et al. proposed an end-to-end expression-tailored generative adversarial network (ET-GAN) to generate talking face videos of identity with enriched facial expression [145]. Expressional video of that identity, instead of identity image and audio, was the model input.

The aforementioned works were concerned with affective facial expression generation on the screen, Benson et al. [146], [147], however, developed a Facial Expression Time Petri Net (FETPN) model to display facial expression on an embodied robotic face. They treated the expression of an affective state as a time-constrained behavior of facial physiognomy and regarded the facial physiognomic features as components of a concurrent system. Discrete facial expression was realized through the movement of critical facial points on latex-made mask (with wires attached to apply pressure and pull/ release the facial mask segments). Huang et al. [148] proposed a facial motion imitation method to transfer facial geometric characteristics from humans to robots.

Affective speech-driven facial animation has also been investigated several times:

Sadiq and Erzin [149] investigated this problem by domain adaption: affective and neutral speech representations were first mapped to a common latent space with smaller bias, then the domain adaption augmented affective representation for each discrete emotion state. An emotion-dependent deep audio-to-visual model was then trained on a public dataset for affective facial expression generation. Both objective and subjective measurements justified the model's performance. Karras et al. [150] developed low latency, audio-driven facial animation models. The end-to-end neural net learned a mapping from input audio to facial coordinates and a latent code that could be used as an intuitive controller of the emotional state of synthesized face. Their model consisted of three modules: formant analysis network, articulation network, and output network. Their model could yield reasonable results for inputs of other speakers, even being trained only on audios from a single speaker. Pham et al. [151] proposed training a LSTM network on a large audio-visual data corpus for real-time facial animation, with audio stream as input. The proposed network could estimate head rotation and facial action unit activations from the speaker's speech. In [152], both acoustic features and phoneme label features were utilized to generate natural looking, speaker-independent lip animations synchronized with affective speech.

### 3.4 Affective movement generation

Body movement, along with facial expression, are both forms of non-verbal communication. Body movement can be characterized by three dimensions, namely, function, execution, and expression [153]. Affective body movement generation focuses on the aspect of expression, which reflects and influences the affective qualities, such as believability and engagement that the movement is conveying [153], [154]. Body movement generation can be realized through either physics-based approaches, in which parameters were adopted to control specific actions or data-driven machine learning approaches that relied on human actors performing actions as the demo. Most affective movement generation works were concerned with body movement and head movement, with few works focusing on hand gesture synthesis [155].

#### 3.4.1 Affective movement generation of virtual agent

Most works concerning affective movement generation for virtual agents adopted physics-based approaches:

In [156], a general parameterized behavior model was adopted to integrate affect expression with functional behaviors. Models were parameterized by both spatial extent and motion dynamics. The model was applied to coverbal gestures with a NAO robot in order to express mood in storytelling scenarios.

Burton et al. [157] proposed to imbue a given trajectory of robot movement with expressive content, which was sampled from a database of movements with expressive qualities. This method worked to find emotionally similar movements from the database, based on Laban movement analysis [158].

Carreno-Medrano et al. [159], [160] tried to represent affective bodily movement through a low-dimensional parameterization based on the spatio-temporal trajectories of some basic joints in the human body. It was assumed that the low-dimensional parameterization encodes the affective state that could also be mapped to whole-body motions.

Xia et al. [161] proposed a real-time style translation model that automatically transformed unlabeled, heterogeneous motion data into new styles. The styles of the output animation could be blended by blending the parameters of distinctive styles. For example, a "neutral" walk can be translated into an "angry"-"strutting" walk by linearly interpolating two distinctive output animation styles: "angry" and "strutting".

In [162], an inverse kinematics (IK) reconstruction model along with a statistical motion resampling scheme were proposed to synthesize high-dimensional full-body movements from low-dimensional end-effector trajectories. An inverse kinematic controller was defined for each limb movement while the resampling scheme was for end-effector and pelvis trajectories generation, with the constraint of preserving underlying emotional states.

There are plenty of works on data-driven approaches for body movement generation, but only a few considering the impact of affect states [163], [164]. Alemi et al. [153] presented an interactive animated agent model with controllable affective movements. A Factored, Conditional Restricted Boltzman Machine (FCRBM) [165] was adopted to control the valence and arousal of walking movements, with a corpus of recorded actor-performed affective walking. Two actors were recruited to perform 9 different expressive combinations of valence (negative, neutral, positive) and arousal (low, neutral, high).

### 3.4.2 Affective robotic movement generation

Using NAO and Keepon [166], Knight and Simmons showed that simple robots could convey complex emotion through motion and varying robot task motions was sufficient to communicate a variety of expressive states [167].

In [168], in order to let robots expressing dominance, authors developed a parameter-based model for head tilt and body expansiveness. The model was applied to a collection of behaviors, which were evaluated by human observers.

Some works used learning from demonstration or imitation learning to let robots learn from human gestures to express emotions [169]: Suguitan et al. [170] investigated robot movement generation through learning from demonstration of non-professional actors, with the help of CycleGANs [171]. The CycleGAN consisted of a forward cycle and a backward cycle. In the forward cycle, an encoder-decoder module was adopted to generate a robot movement with human movement as input. In the backward cycle, the generated robot movement was passed into another encoder-decoder module for human movement reconstruction.

Affective body movement generation does not only exist for virtual agents and embodied humanoid robots. Non-humanoid robots can also express emotion [175]:

Cauchard et al. [172] defined a range of personality traits and emotional attributes that could be encoded in drones through their flight paths and speed. Their results indicated that adding an emotional component was part of the key to success in drones' acceptability. Jørgensen tried expressive movement generation in a soft robot [173]. Rincon et al. proposed an adaptive fuzzy mechanism to adjust the perceived PAD (pleasure, arousal, and dominance) according to the environmental temperature, humidity, luminosity, and human proximity. The PAD values were then used to change the motion trajectory of a robot arm to express different emotional states. The motion trajectory was commanded by the Robust Generalized Predictive Controllers (RGPC) using convex optimization by Youla parametrization. Sial et al. [174] introduced a non-verbal and non-facial method for a "mechanoid robot" to express emotions through gestures. Their results indicated that the motion parameters of robots were linked with the change of emotions. How emotions could be expressed in swarms of miniature mobile robots were investigated in [175].

Claret et al. [176] studied the problem of a robot executing a primary task and simultaneously conveying emotions using body motions, which was defined as a lower priority task. They explored the possibility of using kinematic redundancy of a robot to convey emotions. Happiness and sadness were shown as very well delivered, and calm was moderately conveyed, while fear was not well delivered.

Löffler et al. [177] investigated using output modalities of color, motion, and sound for expressing joy, sadness, fear, and anger in an appearance-constrained social robot. The best expressions for each modality and emotion were selected and systematically combined.

## 4 CHALLENGES

Challenges and potential contributions of existing affective generation models are discussed in this section.

### 4.1 Contextualization

Context, including both cultural context and social context, is a critical problem in delivering affective messages [178].

It has been shown the importance of context in understanding the meaning of words [179] and coherence of sentences [180]. However, among all the reviewed affective text generation works, only Affect-LM [85] considered the context words, although Affect-LM only considered words within the range of a sentence. Among the reviewed works on affective speech synthesis, the work in [181] adopted filters to extract contextual information. So far, we have not found any works considering context in affective facial expression and affective movement generation.

The future works on affective generation models are suggested to take into account the contextual information, which should not be limited within the range of words, sentences, but also the social attitude and cultural preference. Social attitude in dialogues can permeate through the whole interaction process, rather than occur at a certain moment [49]. Moreover, better contextualization can provide more information on emotion boundaries. Existing works assume emotions are either instantaneous or over time, which is not necessarily valid [6].

### 4.2 Offline generation vs. online generation

The exisiting works were all offline generation models that were trained on a large corpus of data simultaneously. Offline learning is known for suffering from the restriction of large memory and slow model updates. In the real-world human machine interaction scenarios, it is expected that plenty of sensors will be deployed to collect a large amount of real-time and personalized data. The data processing speed will influence how fast and efficient an agent can respond to the sudden changes in the environment.

An online learning model treats input data as a running stream and makes small incremental updates to the model.

Online learning models do not need large capacity memory to store input data for training, and it can better fit the latest trends of patterns in the data stream as the influence of past data may be gradually discounted. Affective generative models that update online will likely provide more adaptive and individualized solutions.

### 4.3 Multimodality

In affevtive recognition works, multimodal information are fused to promote the recognition capability as different modalities can reinforce or complement each other. Whereas in multimodal affective generation, different modalities are expected to be generated in a consistent and coherent manner. For example, if well organized, the generated affective text, music and video can provide a comprehensive and immersive environment as a better delivery, as oopposed to that of single modality.

While there have been extensive works on multimodal affective recognition, limited effort has been dedicated for multimodal affective generation. The coordination and synchronization among different modalities have been shown to effectively improve the delivery quality of the generated content as well as the user satisfaction [106]. Future works are encouraged to generate an immersive environment where multiple modalities complement each other to affect users' emotional states.

### 4.4 Affective generation for special groups

Numerous technological systems, including affective recognition ones, have been specifically designed for the welfare and rehabilitation of special groups [182], [183], [184]. In contrast, there is a lack of such consideration in affective generation works.

When designing models to achieve optimal rehabilitation purposes, affective generation should target not only general audiences and users, but also special groups, such as the elderly group with dementia [185], children with developmental disorders like Autism Spectrum Disorder [186], and people with depression [187] and anxiety [188], who need taking special care on emotional support.

For example, individuals with ASD are characterized with the difficulties of perceiving other's emotional cues. As a consequence, affective content generated by an automated agent designed for typically developing individuals may have unwanted or unpredictable emotional influences on individuals with ASD.

### 4.5 Combination of affective generation and affective recognition

Affective recognition and affective generation have been studied separately in most scenarios. However, an intelligent agent should be endowed with the capability of both input signal processing and analysis (affective recognition) and generating appropriate messages as a response (affective generation). Heretofore, we have not found any work combining affective generation with affective recognition when designing an intelligent affective agent.

## 5 CONCLUSIONS

We reviewed affective generation models since 2015, the year when deep learning drew public attention. Affective generation is a technique that aims to generate messages to influence the emotional states of users. Topics including affective generation of text, audio, facial expression, and movement are reviewed. Challenges and potential improvements of existing models are discussed at last. We hope this work can pave the way for future works on developing novel affective generation models.

### ACKNOWLEDGMENT

This work is supported partially by the National Natural Science Foundation of China, No.: 62002090.